\documentclass[10pt,twocolumn,letterpaper]{article}

\usepackage{cvpr}
\usepackage{times}
\usepackage{epsfig}
\usepackage{graphicx}
\usepackage{amsmath}
\usepackage{amssymb}

\usepackage{bbm}

\usepackage{booktabs}  
\usepackage{threeparttable}  
\usepackage{multirow}
\usepackage[breaklinks=true,bookmarks=false]{hyperref}

\cvprfinalcopy 

\def\cvprPaperID{5285} 

\ifcvprfinal\fi

\begin{document}

\title{P2SGrad: Refined Gradients for Optimizing Deep Face Models}

\author{Xiao Zhang$^{1}$\quad Rui Zhao$^{2}$\quad Junjie Yan$^{2}$\quad Mengya Gao$^{2}$\quad Yu Qiao$^{3}$\quad Xiaogang Wang$^{1}$\quad Hongsheng Li$^{1}$\\
$^1$CUHK-SenseTime Joint Laboratory, The Chinese University of Hong Kong \quad
$^2$SenseTime Research \\
$^3$SIAT-SenseTime Joint Lab, Shenzhen Institutes of Advanced Technology, Chinese Academy of Sciences\\
{\tt\small zhangx9411@gmail.com\quad hsli@ee.cuhk.edu.hk}
}

\maketitle
\thispagestyle{empty}

\begin{abstract}

Cosine-based softmax losses~\cite{L2-softmax,CosFace,AM-softmax,ArcFace} significantly improve the performance of deep face recognition networks. However, these losses always include sensitive hyper-parameters which can make training process unstable, and it is very tricky to set suitable hyper parameters for a specific dataset. This paper addresses this challenge by directly designing the gradients for adaptively training deep neural networks. We first investigate and unify previous cosine softmax losses by analyzing their gradients. This unified view inspires us to propose a novel gradient called P2SGrad (Probability-to-Similarity Gradient), which leverages a cosine similarity instead of classification probability to directly update the testing metrics for updating neural network parameters. P2SGrad is adaptive and hyper-parameter free, which makes the training process more efficient and faster. We evaluate our P2SGrad on three face recognition benchmarks, LFW~\cite{LFW}, MegaFace~\cite{MegaFace2}, and IJB-C~\cite{ijbc}. The results show that P2SGrad is stable in training, robust to noise, and achieves state-of-the-art performance on all the three benchmarks.
\end{abstract}


\section{Introduction}

\begin{figure*}
\begin{center}
   \includegraphics[width=0.9\linewidth]{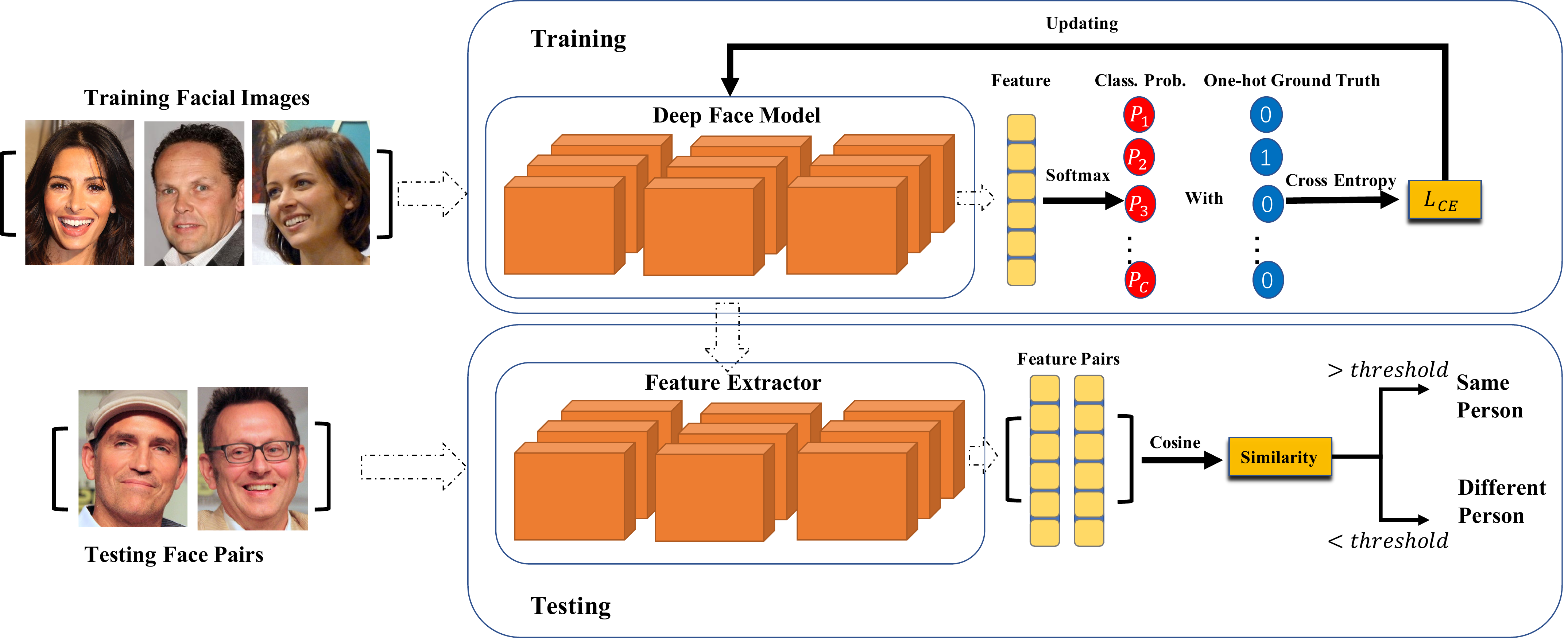}
\end{center}
   \caption{ 
        Pipeline of current face recognition systems. In this general pipeline, deep face models trained on classification tasks are treated as feature extractors. Pairwise similarities between pairs of test images are calculated to determine whether they belong to the same persons. Best viewed in color.
    }
\label{fig:pipeline}

\end{figure*}

Over the last few years, deep convolutional neural networks have significantly boosted the face recognition accuracy. State-of-the-art approaches are based on deep neural networks and adopt the following pipeline: training a classification model with different types of softmax losses and use the trained model as a feature extractor to encode unseen samples. Then the cosine similarities between testing faces' features, are exploited to determine whether these features belong to the same identity. Unlike other vision tasks, such as object detection, where training and testing have the same objectives and evaluation procedures, face recognition systems were trained with softmax losses but tested with cosine similarities. In other words, there is a gap between the softmax probabilities in training and inner product similarities in testing.

This problem is not well addressed in exsiting face recognition models with softmax cross-entropy loss function (softmax loss for short in the remaining part), which mainly considers probability distributions of training classes and ignores the testing setup. In order to bridge this gap, cosine softmax losses~\cite{NormFace,liu_2017_coco_v1,liu_2017_coco_v2} and their angular margin based variants~\cite{CosFace,AM-softmax,ArcFace} directly use cosine distances instead of inner products as the input raw classification scores, namely logits. Specially, the angular margin based variants aim to learn the decision boundaries with a margin between different classes. These methods improve the face recognition performance in the challenging setup.

In spite of their successes, cosine-based softmax loss is only a trade-off: the supervision signals for training are still classification probabilities, which are never evaluated during testing. Considering the fact that the similarity between two testing face images is only related to themselves while the classification probabilities are related to all the identities, cosine softmax losses are not the ideal training measures in face recognition.

This paper aims to address these problems from a different perspective. Deep neural networks are generally trained with gradient-based optimization algorithms where gradients play an essential role in this process. In addition to the loss function, we focus on the gradients of cosine softmax loss functions. This new perspective not only allows us to analyze the relations and problems of previous methods, but also inspires us to develop a novel form of adaptive gradients, P2SGrad, which mitigates the problem of training-testing mismatch and improves the face recognition performance in practice.

To be more specific, P2SGrad optimizes deep models by directly designing new gradients instead of new loss functions. Compared with the conventional gradients in cosine-based softmax losses, P2SGrad uses cosine distances to replace the classification probabilities in the original gradients. P2SGrad also eliminates the effects of different from hyperparameters and the number of classes, and matches testing targets.

This paper mainly contributes in the following aspects:
\begin{enumerate}
    \item We analyze the recent cosine softmax losses and their angular-margin based variants from the perspective of gradients, and propose a general formulation to unify different cosine softmax cross-entropy losses;
    \item  With this unified model, we propose an adaptive hyperparameter-free gradients - P2SGrad, instead of a new loss function for training deep face recognition networks. This method reserves the advantages of using cosine distances in training and replaces classification probabilities with cosine similarities in the backward propagation;
    \item We conduct extensive experiments on large-scale face datasets. Experimental results show that P2SGrad outperforms state-of-the-art methods on the same setup and clearly improves the stability of the training process.
\end{enumerate}

\section{Related Works}
 The accuracy improvements of face recognition~\cite{krizhevsky2012imagenet,hu2017squeeze,parkhi2015deep,szegedy2015going} enjoy the large-scale training data, and the improvements of neural network structures. Modern face datasets contain a huge number of identities, such as LFW~\cite{LFW}, PubFig~\cite{Pubfig}, CASIA-WebFace~\cite{WebFace}, MS1M~\cite{MS-Celeb-1M} and MegaFace~\cite{MegaFace1,MegaFace2}, which enable the effective training of very deep neural networks. A number of recent studies demonstrated that well-designed network architectures lead to better performance, such as DeepFace~\cite{DeepFace}, DeepID2, 3~\cite{DeepID2,DeepID3} and FaceNet~\cite{FaceNet}.

In face recognition, feature representation normalization, which restricts features to lie on a fixed-radius hyper-sphere, is a common operation to enhance models' final performance. COCO loss \cite{liu_2017_coco_v1,liu_2017_coco_v2} and NormFace \cite{NormFace} studied the effect of normalization through mathematical analysis and proposed two strategies through reformulating softmax loss and metric learning. Coincidentally, L2-softmax \cite{L2-softmax} also proposed a similar method. These methods obtain the same formulation of cosine softmax loss from different views.

Optimizing auxiliary metric loss function is also a popular choice for boosting performance. In the early years, most face recognition approaches utilized metric loss functions, such as triplet loss \cite{tripletloss2} and contrastive loss \cite{contrastiveloss}, which use Euclidean margin to measure distance between features. Taking advantages of these works, center loss \cite{centerloss} and range loss \cite{rangeloss} were proposed to reduce intra-class variations through minimizing distance within target classes \cite{belhumeur1997eigenfaces}.

Simply using Euclidean distance or Euclidean margin is insufficient to maximize the classification performance. To circumvent this difficulty, angular margin based softmax loss functions were proposed and became popular in face recognition. Angular constraints were added to traditional softmax loss function to improve feature discriminativeness in L-softmax \cite{L-softmax} and A-softmax \cite{A-softmax}, where A-softmax applied weight normalization but L-softmax \cite{L-softmax} did not. CosFace \cite{CosFace}, AM-softmax \cite{AM-softmax} and ArcFace \cite{ArcFace} also embraced the idea of angular margins and employed simpler as well as more intuitive loss functions compared with aforementioned methods. Normalization is applied to both features and weights in these methods.

\section{Limitations of cosine softmax losses}
In this section we discuss limitations caused by the mismatch between training and testing of face recognition models. We first provide a brief review of the workflow of cosine softmax losses. Then we will reveal the limitations of existing loss functions in face recognition from the perspective of forward and backward calculation respectively.

\subsection{Gradients of cosine softmax losses}

In face recognition tasks, the cosine softmax cross-entropy loss has an elegant two-part formulation, softmax function and cross-entropy loss. 

We discuss softmax function at first. Assuming that the vector $\vec x_i$ denotes the feature representation of a face image, the input of the softmax function is the logit $f_{i,j}$, \ie,
\begin{equation}
f_{i,j}=s\cdot{\frac{\langle \vec x_{i},\vec W_{j} \rangle}{\|\vec x_i\|_2\|\vec W_j\|_2}}=s\cdot\langle\hat{\textbf{x}}_i, \hat{\textbf{W}}_j\rangle=s\cdot\cos{\theta_{i,j}}
\text{, }
\label{eq:f_cosinesoftmax}
\end{equation}
where $s$ is a hyperparameter and $f_{i,j}$ is the classification score (logit) that $\vec x_i$ is assigned to class $j$, and $W_j$ is the weight vector of class $j$. $\hat{\textbf{x}}_i$ and $\hat{\textbf{W}}_j$ are normalized vectors of $x_i$ and $W_j$ respectively. $\theta_{i,j}$ is the angle between feature $x_i$ and class weight $W_j$. 
The logits $f_{i,j}$ are then input into the softmax function to obtain the probability $P_{i,j}=\text{Softmax}(f_{i,j})=\frac{e^{f_{i,j}}}{\sum_{k=1}^{C}e^{f_{i,k}}}$, where $C$ is the number of classes and the output $P_{i,j}$ can be interpreted as the probability of $\vec x_i$ being assigned to a certain class $j$. If $j=y_i$, then $P_{i,y_i}$ is the class probability of $\vec x_i$ being assigned to its corresponding class $y_i$.

Then we discuss the cross-entropy loss associated with the softmax function, which measures the divergence between the predicted probability $P_{i,y_i}$ and ground truth distributions as
\begin{equation}
\mathcal L_{\text{CE}}(\vec x_i)=-\log{P_{i,y_i}}=-\log{\frac{e^{f_{i,y_i}}}{\sum_{k=1}^{C}e^{f_{i,k}}}}
\text{, }
\label{eq:CE_loss}
\end{equation}
where $\mathcal L_{\text{CE}}(\vec x_i)$ is the loss of input feature $\vec x_i$. The larger probability $P_{i,y_i}$ is, the smaller loss $\mathcal L_{\text{CE}}(\vec x_i)$ is.

In order to decrease the loss $\mathcal L_{\text{CE}}(\vec x_i)$, the model needs to enlarge $P_{i,y_i}$ and thus enlarges $f_{i,y_i}$. Then $\theta_{i,y_i}$ becomes smaller. In summary, cosine softmax loss function maps $\theta_{i,y_i}$ to the probability $P_{i,y_i}$ and calculates the cross-entropy loss to supervise the training. 

\begin{figure}
\centering
\includegraphics[width=0.7\linewidth]{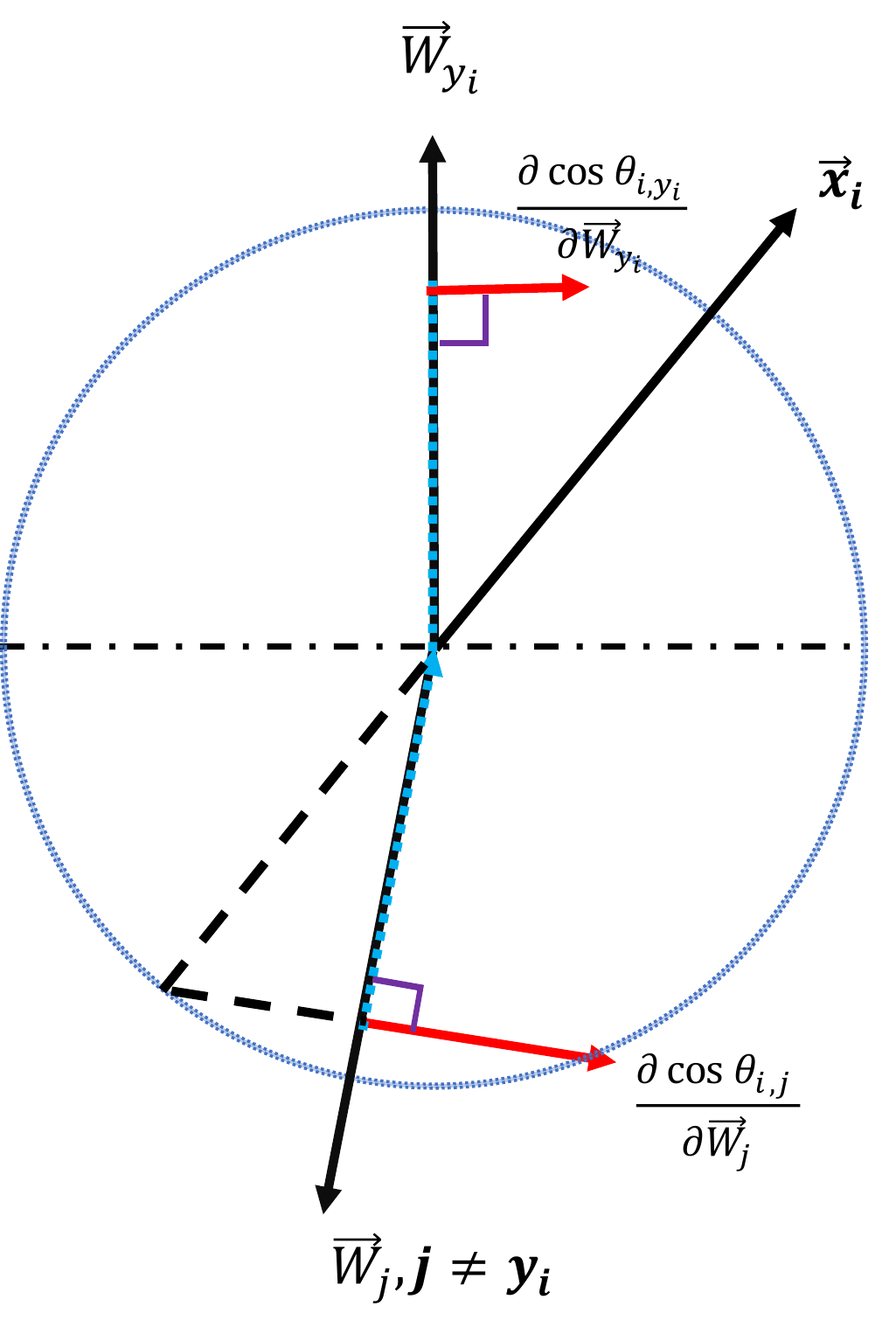}
\caption{ Gradient direction of $\frac{\partial \cos{\theta_{i,j}}}{\partial {\Vec{W}_j}}$. Note this gradient is the updating direction of $\vec{W}_j$. The red pointed line shows that the gradient of $\vec{W}_j$ is vertical to $\vec{W}_j$ itself and in the plane spanned by $\vec{x}_i$ and $\vec{W}_j$. This can be seen as the fastest direction for updating $\vec W_{y_i}$ to be close to $\vec{x}_i$ and for updating $\vec{W}_j, j\neq{y_i}$ to be far away from $\vec x_i$. Best viewed in color.}
\label{fig:cos_w_grad}

\end{figure}

In the backward propagation process, classification probabilities $P_{i,j}$ play key roles in optimization. The gradient of $\vec x_i$ and $\vec W_j$ in cosine softmax losses are calculated as
\begin{equation}
\begin{aligned}
\frac{\partial\mathcal{L}_{\text{CE}}(\vec x_i)}{\partial \Vec{x}_i}&=\sum^{C}_{j=1}(P_{i,j} -\mathbbm{1}( y_i =j))
\nabla{f(\cos{\theta_{i,j}})}\cdot\frac{\partial \cos{\theta_{i,j}}}{\partial {\Vec{x}_i}},\\
\frac{\partial\mathcal{L}_{\text{CE}}(\vec x_i)}{\partial \Vec{W}_j}&=(P_{i,j} - \mathbbm{1}( y_i =j))\nabla{f(\cos{\theta_{i,j}})}\cdot\frac{\partial \cos{\theta_{i,j}}}{\partial {\Vec{W}_j}}
\text{, }
\label{eq:gradient_x_w}
\end{aligned}
\end{equation}
where the indicator function $\mathbbm{1}(j=y_i)$ returns $1$ when $j = y_i$ and $0$ otherwise.
$\frac{\partial \cos{\theta_{i,j}}}{\partial {\Vec{x}_i}}$ and $\frac{\partial \cos{\theta_{i,j}}}{\partial {\Vec{W}_j}}$ can be computed respectively as:
\begin{equation}
\begin{aligned}
\frac{\partial \cos{\theta_{i,j}}}{\partial {\Vec{x}_i}}&=\frac{1}{\|\vec x_i\|_2}(\hat{\textbf{W}}_j-\cos{\theta_{i,j}}\cdot{\hat{\textbf{x}}_i}),\\
\frac{\partial \cos{\theta_{i,j}}}{\partial {\Vec{W}_j}}&=\frac{1}{\|\vec W_j\|_2}(\hat{\textbf{x}}_i-\cos{\theta_{i,j}}\cdot{\hat{\textbf{W}}_j})
\text{, }
\label{eq:cos_xw_grad}
\end{aligned}
\end{equation}
where $\hat{\textbf{W}}_j$ and $\hat{\textbf{x}}_i$ are unit vectors of $\vec W_j$ and $\vec x_i$, respectively. $\frac{\partial \cos{\theta_{i,j}}}{\partial {\Vec{W}_j}}$ are visualized as the red arrows in Fig.~\ref{fig:cos_w_grad}. This gradient vector is the updating directions of class weights $\vec W_j$. Intuitively, we expect the updating of $\vec W_j$ makes $\vec W_{y_i}$ close to $\vec x_i$, and makes $\vec W_j$ for $j\neq{y_i}$ away from $\vec x_i$. Gradient $\frac{\partial \cos{\theta_{i,j}}}{\partial {\Vec{W}_j}}$ is vertical to $\vec W_j$ and points toward $\vec x_i$. Thus it is the fastest and optimal directions for updating 
$\vec W_j$.

Then we consider the gradient $\nabla{f(\cos{\theta_{i,j}})}$. In conventional cosine softmax losses~\cite{L2-softmax,NormFace,liu_2017_coco_v1}, the classification score $f(\cos{\theta_{i,j}})=s\cdot{\cos{\theta_{i,j}}}$ and thus $\nabla{f(\cos{\theta_{i,j}})} = s$. In angular margin-based cosine softmax losses~\cite{AM-softmax,CosFace,ArcFace}, however, the gradient of $f_{\text{margin}}(\cos{\theta_{i,y_i}})$ for $j=y_i$ depends on where the margin parameter $m$ is. In CosFace~\cite{CosFace} $f(\cos{\theta_{i,y_i}})=s\cdot(\cos{\theta_{i,y_i}-m)}$, thus $\nabla{f(\cos{\theta_{i,y_i}})}=s$ and in ArcFace~\cite{ArcFace} $f(\cos{\theta_{i,y_i}})=s\cdot\cos{(\theta_{i,y_i}+m)}$, thus $\nabla{f(\cos{\theta_{i,y_i}})}=s\cdot\frac{\sin{(\theta_{i,y_i}+m)}}{\sin{\theta_{i,y_i}}}$.
In general, gradient $\nabla{f(\cos{\theta_{i,j}})}$ is always a scalar related to parameters $s$, $m$ and $\cos{\theta_{i,j}}$.

Based on the aforementioned discussions, we reconsider gradients of class weights $\vec W_j$ in Eq.~\eqref{eq:gradient_x_w}. In $\frac{\partial\mathcal{L}_\text{CE}}{\partial \Vec{W}_j}$, the first part $(P_{i,j} - \mathbbm{1}( y_i =j)\cdot\nabla{f(\cos{\theta_{i,j}})}$ is a scalar, which decides the length of gradient, while the second part $\frac{\partial \cos{\theta_{i,j}}}{\partial {\Vec{W}_j}}$ is a vector which decides the direction of gradients. \emph{Since the directions of gradients for various cosine softmax losses remain the same, the essential difference of these cosine softmax losses is the different lengths of gradients, which significantly affect the optimization of model.} In the following sections, we will discuss the suboptimal gradient length caused by forward and backward process respectively.

\subsection{Limitations in probability calculation}

In this section we discuss the limitations of the forward calculation of cosine softmax losses in deep face networks and focus on the classification probability $P_{i,j}$ obtained in the forward calculation.

We first revisit the relation between $P_{i,j}$ and $\theta_{i,j}$. 
The classification probability $P_{i,j}$ in Eq.~\eqref{eq:gradient_x_w} is a part of gradient length. Hence $P_{i,j}$ significantly affects the length of gradient. 
Probability $P_{i,j}$ and logit $f_{i,j}$ are positively correlated. For all cosine softmax losses, logits $f_{i,j}$ measure $\theta_{i,j}$ between feature $\vec x_i$ and class weight $\vec W_j$. A larger $\theta_{i,j}$ produces lower classification probability $P_{i,j}$ while a smaller $\theta_{i,j}$ produces higher $P_{i,j}$. It means that $\theta_{i,j}$ affects gradient length by its corresponding probability $P_{i,j}$. 
The equation sets up a mapping relation between $\theta_{i,j}$ and $P_{i,j}$ and makes $\theta_{i,j}$ affects optimization. Above analysis is also the reason why cosine softmax losses are effective in face recognition performance.

Since $\theta_{i,y_i}$ is the direct measurement of the generalization but it can only indirectly affect gradients by corresponding $P_{i,y_i}$, setting a reasonable mapping relation between $\theta_{i,y_i}$ and $P_{i,y_i}$ is crucial. However, there are two tricky problems in current cosine softmax losses: (1) classification probability $P_{i,y_i}$ is sensitive to hyperparameter settings; (2) the calculation of $P_{i,y_i}$ is dependent on class number, which is not related to face recognition tasks. We will discuss these problems below.

\begin{figure}[t]
\begin{center}
   \includegraphics[width=1\linewidth]{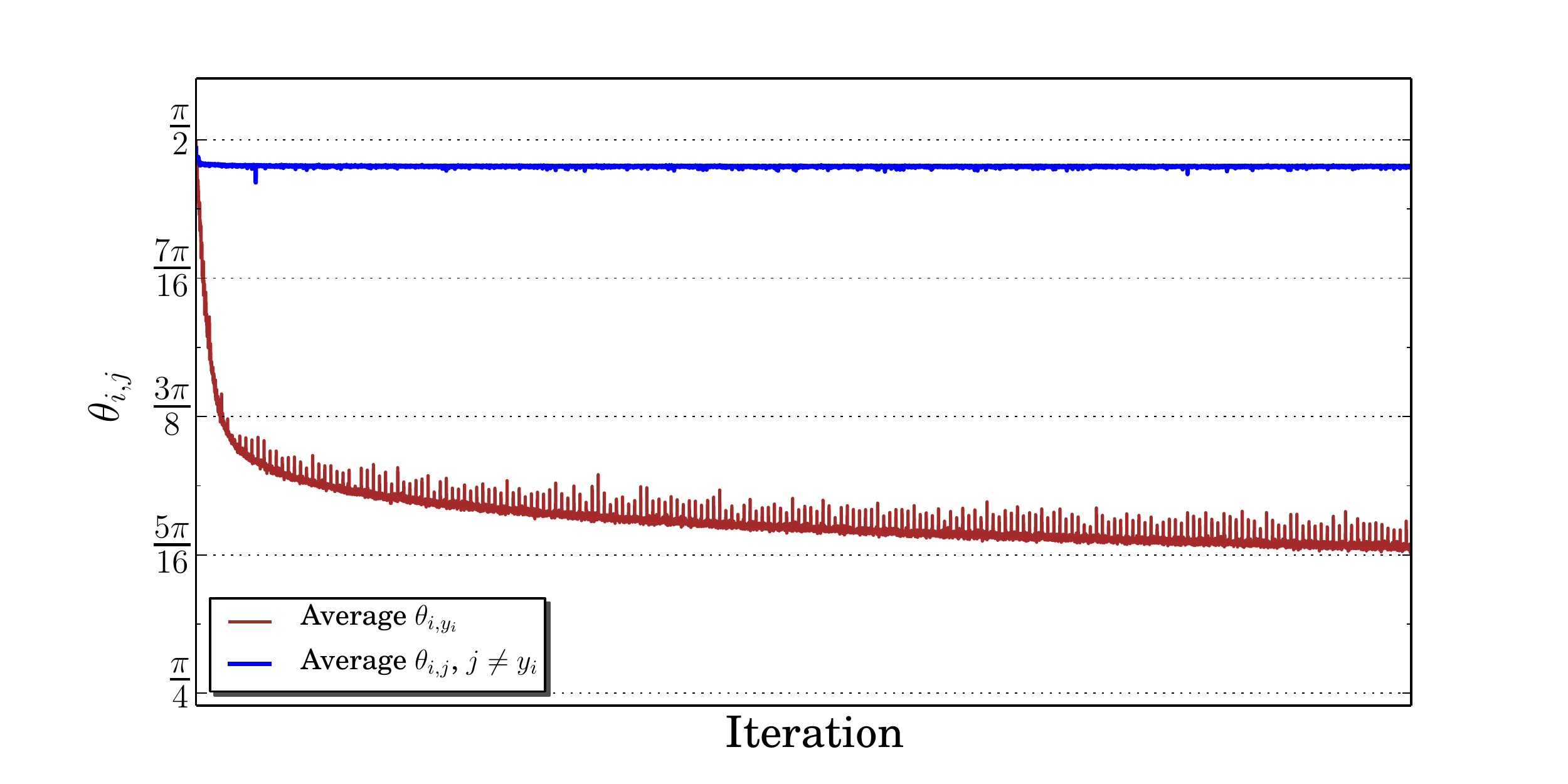}
\end{center}
   \caption{ 
        The change of average $\theta_{i,j}$ of each mini-batch when training on WebFace dataset. (Red) average angles in each mini-batch for non-corresponding classes, $\theta_{i,j}$ for $j \neq y_i$. (Brown) average angles in each mini-batch for corresponding classes, $\theta_{i,y_i}$.}
\label{fig:theat_range}

\end{figure}

{\bf $P_{i,y_i}$ is sensitive to hyperparameters.}
The most common hyperparameters in conventional cosine softmax losses~\cite{L2-softmax,NormFace,liu_2017_coco_v1} and margin variants~\cite{ArcFace} are the scale parameter $s$ and the angular margin parameter $m$.
We will analyze the sensitivity of probability $P_{i,y_i}$ to hyperparameter $s$ and $m$. For more accurate analysis, we first look at the actual range of $\theta_{i,j}$.
Fig.~\ref{fig:theat_range} exhibits how the average $\theta_{i,j}$ changes in training. Mathematically, $\theta_{i,j}$ could be any value in $[0, \pi]$. In practice, however, the maximum $\theta_{i,j}$ is around $\frac{\pi}{2}$. The blue curve reveals that $\theta_{i,j}$ for $j\neq{y_i}$ do not change significantly during training. The brown curve reveals that $\theta_{i,y_i}$ is gradually reduced. Therefore we can reasonably assume that $\theta_{i,j}\approx{\frac{\pi}{2}}$ for $j\neq{y_i}$ and the range of $\theta_{i,y_i}$ is $[0,\frac{\pi}{2}]$. Then $P_{i,y_i}$ can be rewritten as
\begin{equation}
\begin{aligned}
P_{i,y_i}& =\frac{e^{f_{i,y_i}}}{\sum_{k=1}^{C}e^{f_{i,k}}} =\frac{e^{f_{i,y_i}}}{e^{f_{i,y_i}}+\sum_{k\neq{y_i}}e^{s\cdot\cos{\theta_{i,k}}}}\\
& \approx \frac{e^{f_{i,y_i}}}{e^{f_{i,y_i}}+\sum_{k\neq{y_i}}e^{s\cdot\cos{\pi/2}}} =  \frac{e^{f_{i,y_i}}}{e^{f_{i,y_i}}+(C-1)},
\end{aligned}
\label{eq:p_i_j_C-1}
\end{equation}
where $f_{i,y_i}$ is logit that $\vec x_i$ is assigned to its corresponding class $y_i$, and $C$ is the class number. 

We can obtain the mapping between probability $P_{i,y_i}$ and angle $\theta_{i,y_i}$ under different hyperparameter settings. In state-of-the-art angular margin based losses~\cite{ArcFace}, logit $f_{i,y_i}=s\cdot\cos{(\theta_{i,y_i}+m)}$.
\begin{figure}[t]
\begin{center}
   \includegraphics[width=1\linewidth]{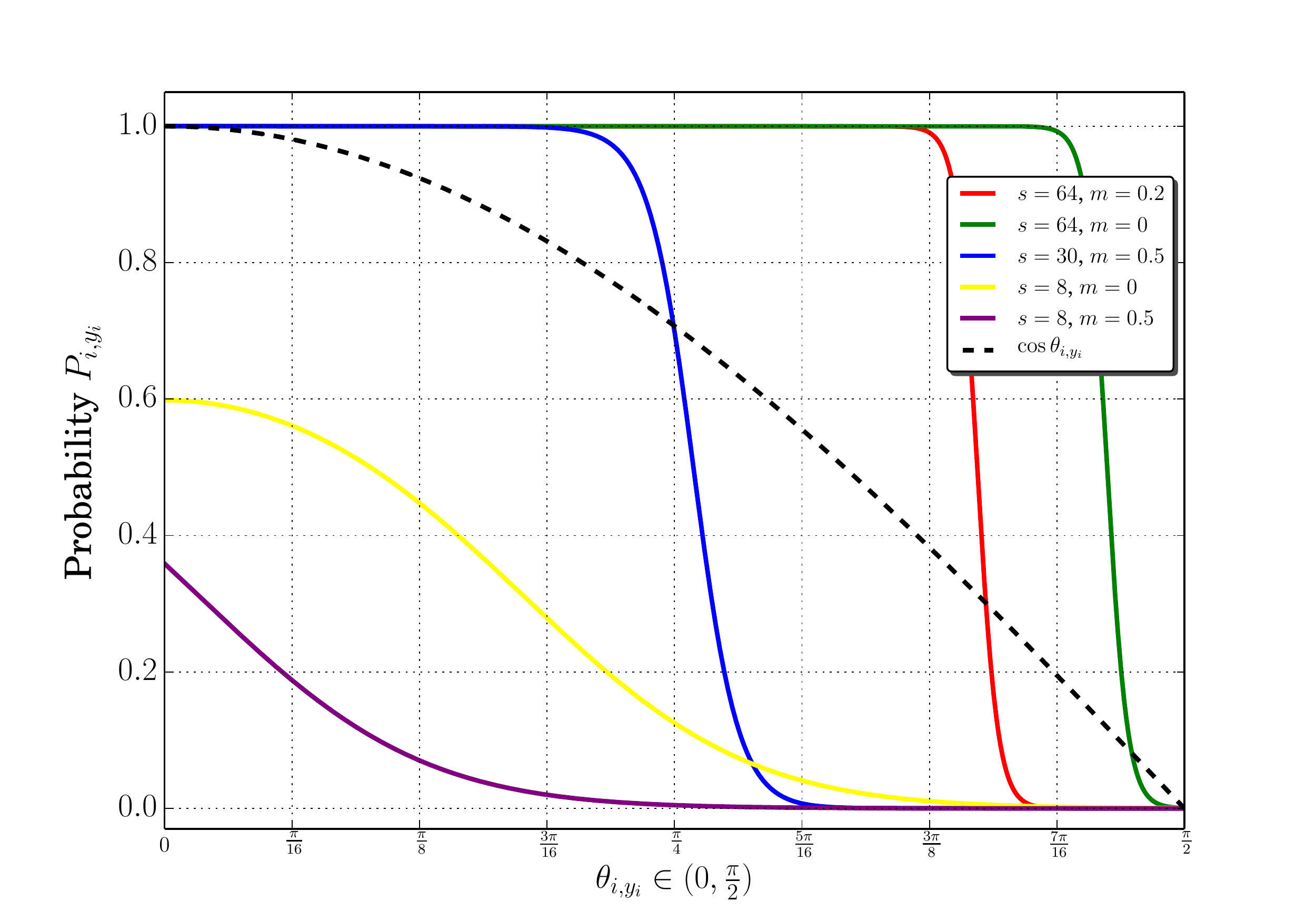}
\end{center}
   \caption{ 
        Probability $P_{i,y_i}$ curves w.r.t. the angle $\theta_{i,y_i}$ with different hyperparameter settings.}
\label{fig:theat_prob}

\end{figure}
Fig.~\ref{fig:theat_prob} reveals that different settings of $s$ and $m$ can significantly affect the relation between $\theta_{i,y_i}$ and $P_{i,y_i}$. Apparently, both the green curve and the purple curve are examples of unreasonable relations. The former is so lenient that even a very larger $\theta_{i,y_i}$ can produce a large $P_{i,y_i}\approx{1}$. The later is so strict that even a very small $\theta_{i,y_i}$ can just produce a low $P_{i,y_i}$. In short, for a specific degree of $\theta_{i,y_i}$, the probabilities $P_{i,y_i}$ under different settings are very different. This observation indicates that probability $P_{i,y_i}$ is sensitive to parameters $s$ and $m$. 

\begin{figure}[t]
\begin{center}
   \includegraphics[width=1\linewidth]{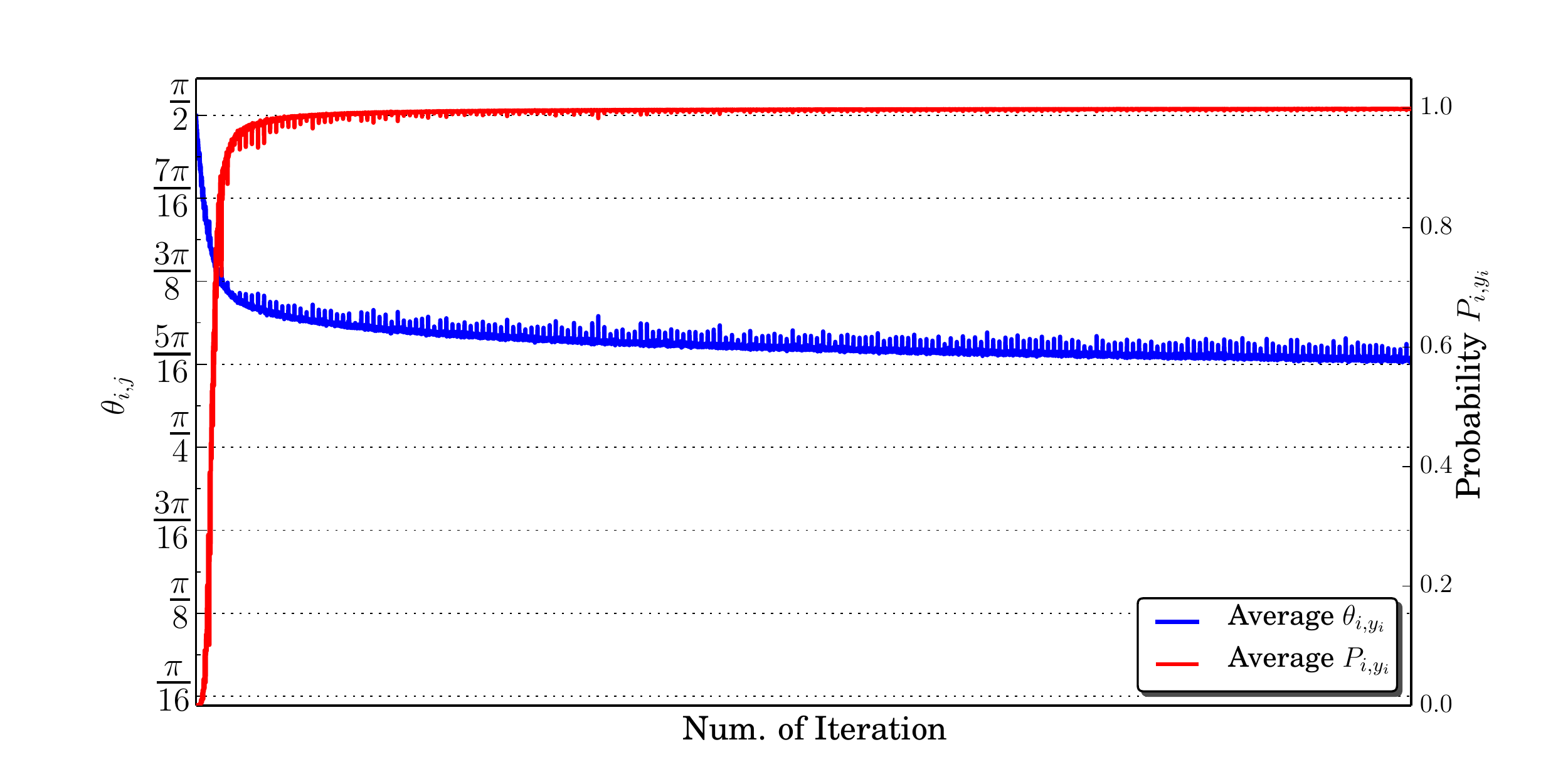}
\end{center}
   \caption{ 
        The change of probability $P_{i,y_i}$ and angle $\theta_{i,y_i}$ as the iteration number increases with the hyperparameter setting $s=35$ and $m=0.2$. Best viewed in color.}
\label{fig:theat_prob_vs_iter}

\end{figure}

To further confirm this conclusion, we take an example of correspondences between $P_{i,y_i}$ and $\theta_{i,y_i}$ in real training.
In Fig.~\ref{fig:theat_prob_vs_iter}, the red curve represents the change of $P_{i,y_i}$ and the blue curve represents the change of $\theta_{i,y_i}$ during the training process. As we discussed above, $P_{i,y_i}\approx{1}$ can produce very short gradients so that the sample $\vec{x}_i$ has little affection in updating. This setting is not ideal because $P_{i,y_i}$ increases to $1$ rapidly but $\theta_{i,y_i}$ is still large.
Therefore classification probability $P_{i,y_i}$ largely depends on the setting of the hyperparameter $s$.

{\bf $P_{i,y_i}$ contains class number.}
In closed-set classification problems, probabilities $P_{i,j}$ become smaller as the growth of class number $C$. This is reasonable in classification tasks.
However, this is not suitable for face recognition, which is an open-set problem. Since $\theta_{i,y_i}$ is the direct measurement of generalization of $\vec x_i$ while $P_{i,y_i}$ is the indirect measurement, we expect that they have a consistent semantic meaning. But $P_{i,y_i}$ is related to class nubmer $C$ while $\theta_{i,y_i}$ is not, which causes the mismatch between them.

\begin{figure}[t]
\begin{center}
   \includegraphics[width=1\linewidth]{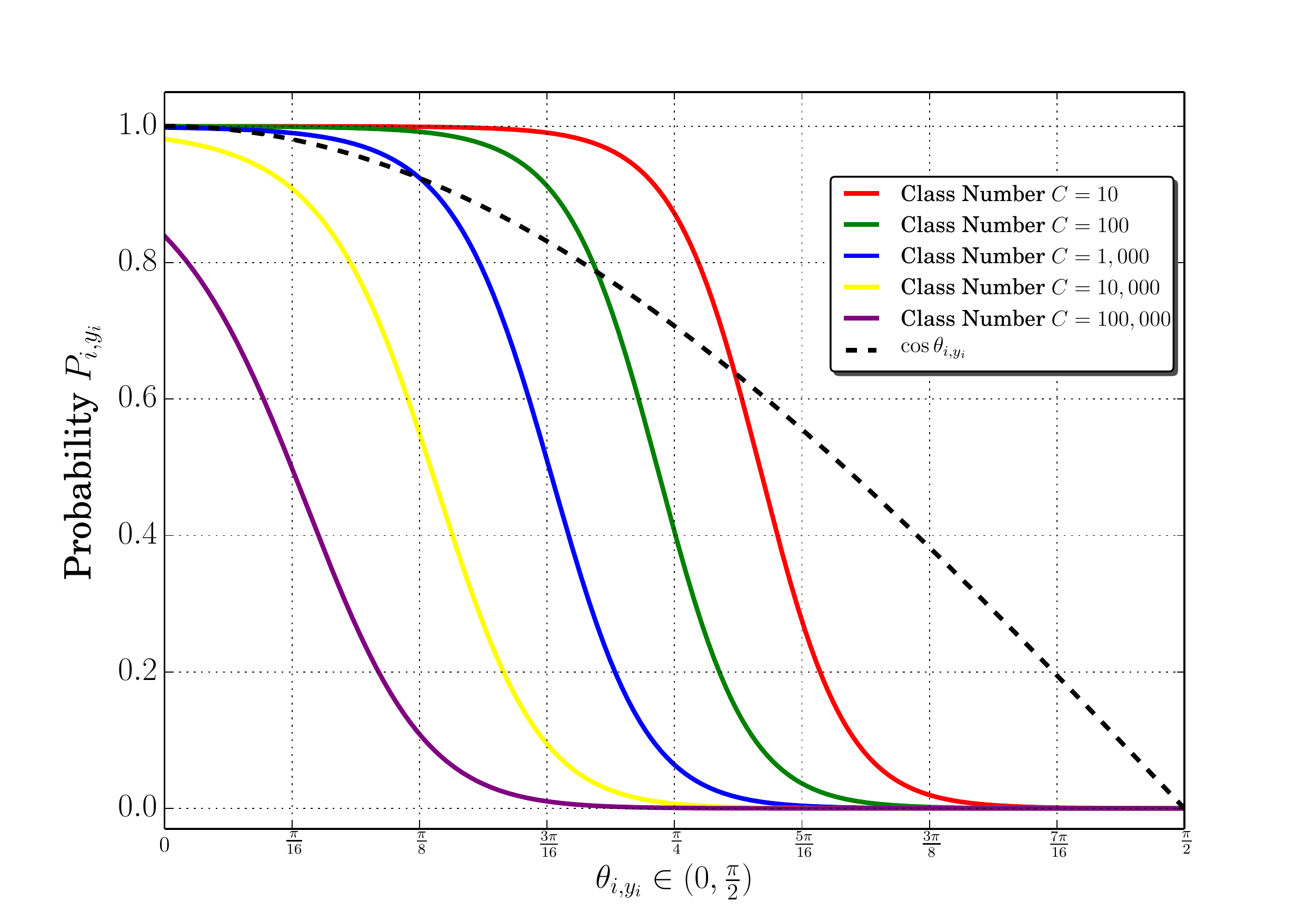}
\end{center}
   \caption{  
        {$P_{i,y_i}$ with different class numbers.} The hyperparameter setting is fixed to $s=15$ and $m=0.5$ for fair comparison. Best viewed in color.}
\label{fig:classnum_prob}

\end{figure}

As shown in Fig.~\ref{fig:classnum_prob}, the class number $C$ is an important factor for $P_{i,y_i}$.

From the above discussion, we reveal that limitations exist in the forward calculation of cosine softmax losses. Both hyperparameters and the class number, which are unrelated to face recognition tasks, can determine the probability $P_{i,y_i}$, and thus affect the gradient length in Eq.~\eqref{eq:gradient_x_w}.

\subsection{Limitation in backward calculation of cosine softmax losses}
In this section, we discuss the limitations in the backward calculation of the cosine softmax function, especially the angular-margin based softmax losses~\cite{ArcFace}.

We revisit the gradient $\nabla{f(\cos{\theta_{i,j}})}$ in Eq.~\eqref{eq:gradient_x_w}. Besides $P_{i,y_i}$, the part of $\nabla{f(\cos{\theta_{i,j}})}$ also affects the length of gradient. Larger $\nabla{f(\cos{\theta_{i,j}})}$ produces longer gradients while smaller ones produce shorter gradients. So we expect $\theta_{i,y_i}$ and values of $\nabla{f(\cos{\theta_{i,j}})}$ to be positively correlated: small $\theta_{i,y_i}$ for small $\nabla{f(\cos{\theta_{i,j}})}$ and large $\theta_{i,y_i}$ for larger $\nabla{f(\cos{\theta_{i,j}})}$.

\begin{figure*}
\begin{center}
   \includegraphics[width=1\linewidth]{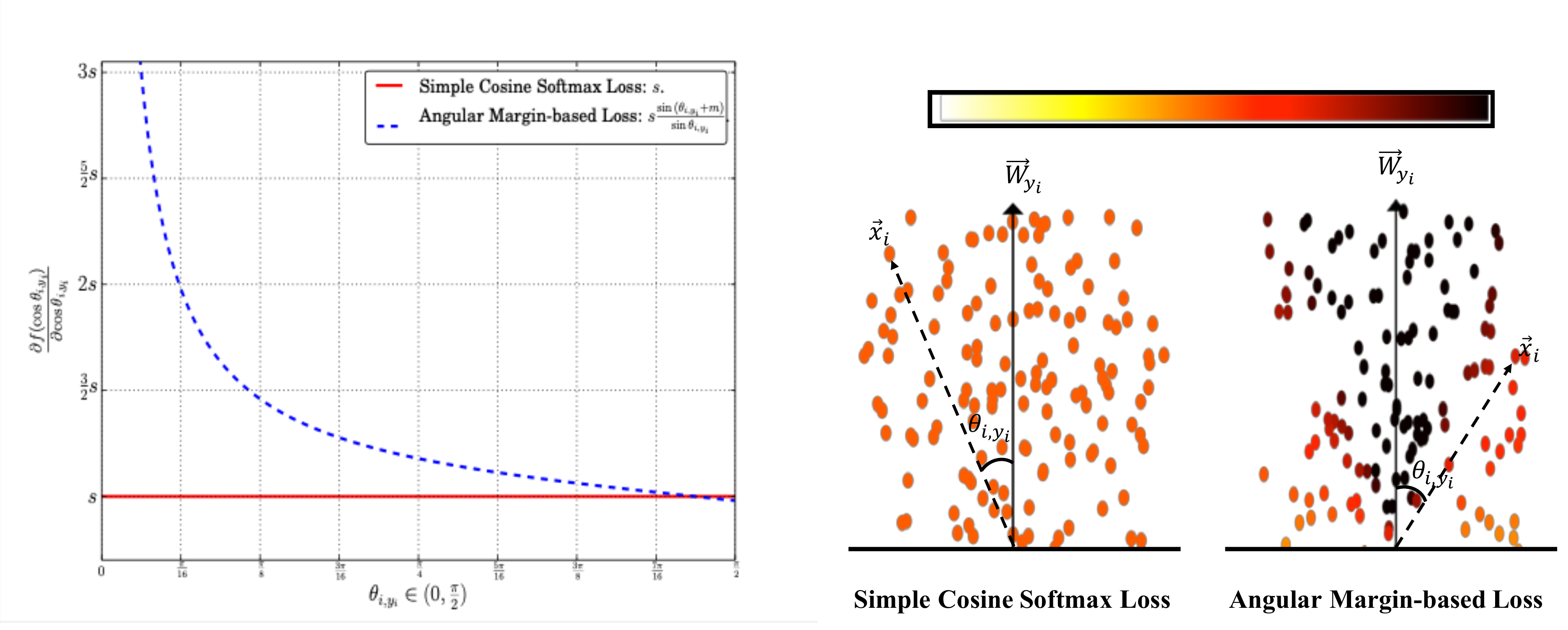}
\end{center}
   \caption{ 
        How $\nabla{f(\cos{\theta_{i,j}})}$ affects the length of gradients. (Left) the correspondence between $\theta_{i,y_i}$ and $\nabla{f(\cos{\theta_{i,j}})}$. The red curve means $\nabla{f(\cos{\theta_{i,j}})}$ is constant in conventional cosine softmax losses~\cite{L2-softmax,NormFace,liu_2017_coco_v1} while the blue curve means small $\theta_{i,y_i}$ can produce very large $\nabla{f(\cos{\theta_{i,j}})}$. (Right) each point refers to a feature $\vec x_i$ and the vertical vector is weight $\vec W_{y_i}$. The $\theta_{i,y_i}$ is angle between each $\vec x_i$ and $\vec W_{y_i}$. The color from light to dark corresponds to the value of $\nabla{f(\cos{\theta_{i,j}})}$ from small to large. Hence for the factor of $\nabla{f(\cos{\theta_{i,j}})}$, the dark points produce longer gradients than the light points. Best viewed in color.}
\label{fig:grad_f_cos}
\end{figure*}

The logit $f_{i,y_i}$ is different in various cosine softmax losses, and thus the specific form of $\nabla{f(\cos{\theta_{i,j}})}$ is different. Generally, we focus on simple cosine softmax losses~\cite{L2-softmax,NormFace,liu_2017_coco_v1} and state-of-the-art angular margin based loss~\cite{ArcFace}.
Their $\nabla{f(\cos{\theta_{i,j}})}$ are visualized in Fig.~\ref{fig:grad_f_cos}, which show that the lengths of gradients in conventional softmax cosine losses~\cite{L2-softmax,NormFace,liu_2017_coco_v1} are constant. However, in angular margin-based losses~\cite{ArcFace}, the lengths of gradients and $\theta_{i,y_i}$ are negatively correlated, which is completely contrary to our expectations. Moreover, the correspondence between length of gradients in angular margin-based loss~\cite{ArcFace} and $\theta_{i,y_i}$ becomes tricky: when $\theta_{i,y_i}$ gradually reduced, $P_{i,y_i}$ tends to shorten length of gradients but $\nabla{f(\cos{\theta_{i,j}})}$ tends to elongate the length. Therefore, the geometric meaning of the gradient length becomes self-contradictory in angular margin-based cosine softmax loss. 

\subsection{Summary}

In the above discussion, we first reveal that various cosine softmax losses have the same updating directions. Hence the main difference between the variants are their gradient lengths. 
For the length of gradient, there are two scalars that determine its value: the probability $P_{i,y_i}$ in the forward process and the gradient $\nabla{f(\cos{\theta_{i,j}})}$. For $P_{i,y_i}$, we observe that it can be substantially affected by different hyperparameter settings and class numbers. For $\nabla{f(\cos{\theta_{i,j}})}$, its value depends on the definition of $f(\cos{\theta_{i,y_i}})$.

In summary, from the perspective of gradient, the widely used cosine softmax losses~\cite{L2-softmax,NormFace,liu_2017_coco_v1} and their angular margin variants~\cite{ArcFace} cannot produce optimal gradient lengths with well-explained geometric meanings.

\section{P2SGrad: Change Probability to Similarity in Gradient}

In this section, we propose a new method, namely P2SGrad, that determines the gradient length only by $\theta_{i,j}$ in training face recognition models. Formally, the gradient length produced by P2SGrad is hyperparameter-free and not related to the number of class $C$ nor to a ad-hoc definition of logit $f_{i,y_i}$. P2SGrad does not need a specified formulation of loss function because gradients is well-designed to optimize deep models.

Since the main differences between state-of-the-art cosine softmax losses are the gradient lengths, reforming a reasonable gradient length is an intuitive thought. In order to decouple the length factor and direction factor of the gradients, we rewrite Eq.~\eqref{eq:gradient_x_w} as
\begin{equation}
\begin{aligned}
\nabla\mathcal{L}_{\text{CE}}(\vec x_i)&=\sum^{C}_{j=1}L(P_{i,j},f(\cos{\theta_{i,j}}))\cdot D(\vec{W}_j,\vec{x}_i),\\
\nabla\mathcal{L}_{\text{CE}}(\vec W_j)&=L(P_{i,j},f(\cos{\theta_{i,j}}))\cdot D(\vec{x}_i,\vec{W}_j)
\text{, }
\label{eq:gradient_x_w_in_length_direction}
\end{aligned}
\end{equation}
where the direction factors $D(\vec{W}_j,\vec{x}_i)$ and $D(\vec{x}_i,\vec{W}_j)$ are defined as 

\begin{equation}
\begin{aligned}
D(\vec{W}_j,\vec{x}_i)&=\frac{1}{\|\vec x_i\|_2}(\hat{\textbf{W}}_j-\cos{\theta_{i,j}}\cdot{\hat{\textbf{x}}_i}),\\
D(\vec{x}_i,\vec{W}_j)&=\frac{1}{\|\vec W_j\|_2}(\hat{\textbf{x}}_i-\cos{\theta_{i,j}}\cdot{\hat{\textbf{W}}_j})
\text{,}
\label{D_gradient}
\end{aligned}
\end{equation}
where $\hat{\textbf{W}}_j$ and $\hat{\textbf{x}}_i$ are unit vectors of $\vec W_j$ and $\vec x_i$, respectively. $\cos{\theta_{i,j}}$ is the cosine distances between feature $\vec x_i$ and class weights $\vec W_j$. The direction factors will not be changed because they are the fastest changing directions, which are specified before.
The length factor $| L(P_{i,j},f(\cos{\theta_{i,j}}))|$ is defined as
\begin{align}
    |L(P_{i,j},f(\cos{\theta_{i,j}}))|
    =\begin{cases}
    (1-P_{i,y_i})|\nabla{f(\cos{\theta_{i,y_i}})}| & j = y_i,\\
    \displaystyle
    P_{i,j}\cdot|\nabla{f(\cos{\theta_{i,j}})}| & j \neq y_i.
    \end{cases}
    \label{L_gradient}
\end{align}
The length factor $|L(P_{i,j},f(\cos{\theta_{i,j}}))|$ depends on the probability $P_{i,j}$ and $\nabla{f(\cos{\theta_{i,j}})}$ which are what we aim to reform.

Since we expect that the new length is hyperparameter-free, the cosine logit $f(\cos{\theta_{i,j}})$ will not have hyperparameters like $s$ or $m$. Thus a constant $\nabla{f(\cos{\theta_{i,j}})}$ should be an ideal choice.

For the probability $P_{i,j}$, because it is hard to set a reasonable mapping function between $\theta_{i,j}$ and $P_{i,j}$, we can directly use $\cos\theta_{i,j}$ as a good alternative of $P_{i,j}$ in the gradient length term. Firstly, they have the same theoretical range of $[0,1]$ where $\theta_{i,j}\in{[0,\frac{\pi}{2}]}$. Secondly, unlike $P_{i,j}$ which is adversely influenced by hyperparameter and the number of class, $\cos{\theta_{i,j}}$ does not contain any of these. It means that we do not need to select specified parameters settings for ideal correspondence between $\theta_{i,y_i}$ and $P_{i,y_i}$. Moreover, compared with $P_{i,j}$, $\cos{\theta_{i,j}}$ is a more natural supervision because cosine similarities are used in the testing phase of open-set face recognition systems while probabilities only apply for close-set classification tasks. Therefore, our reformed gradient length factor $\tilde{L}(\cos{\theta_{i,j}})$ can be defined as:
\begin{align}
    \tilde{L}(\cos{\theta_{i,j}})=\cos{\theta_{i,j}}-\mathbbm{1}(j=y_i),
    \label{L_p2s}
\end{align}
where $\tilde{L}(\cos{\theta_{i,j}})$ is a function of $\cos{\theta_{i,j}}$. The reformed gradients $\tilde{G}_{\text{P2SGrad}}$ could then be defined as
\begin{equation}
\begin{aligned}
&\tilde{G}_{\text{P2SGrad}}(\vec x_i)=\sum^{C}_{j=1}\tilde{L}(\cos{\theta_{i,j}})\cdot D(\vec{W}_j,\vec{x}_i),\\
&\tilde{G}_{\text{P2SGrad}}(\vec W_j)=\tilde{L}(\cos{\theta_{i,j}})\cdot D(\vec{x}_i,\vec{W}_j)
\text{, }
\label{eq:P2SGrad0}
\end{aligned}
\end{equation}
where $\mathbbm{1}$ is the indicator function. The full formulation can be rewrite as
\begin{equation}
\begin{aligned}
&\tilde{G}_{\text{P2SGrad}}(\vec x_i)=\sum^{C}_{j=1}(\cos{\theta_{i,j}}-\mathbbm{1}(j=y_i))\cdot\frac{\partial \cos{\theta_{i,j}}}{\partial {\Vec{x}_i}},\\
&\tilde{G}_{\text{P2SGrad}}(\vec W_j)=(\cos{\theta_{i,j}}-\mathbbm{1}(j=y_i))\cdot\frac{\partial \cos{\theta_{i,j}}}{\partial {\Vec{W}_j}}
\text{, }
\label{eq:P2SGrad}
\end{aligned}
\end{equation}

The formulation of P2SGrad is not only succinct but reasonable. When $j=y_i$, the proposed gradient length and $\theta_{i,j}$ are positively correlated, when $j\neq{y_i}$, they are negatively correlated. More importantly, gradient length in P2SGrad only depends on $\theta_{i,j}$ and thus is consistent the testing metric of face recognition systems.

\section{Experiments}
In this section, we conduct a series of experiments to evaluate the proposed P2SGrad.  We first verify advantages of P2SGrad in some exploratory experiments by testing the model's performance on LFW~\cite{LFW}. Then we evaluate P2SGrad on MegaFace~\cite{MegaFace2} Challenge and IJBC 1:1 verification~\cite{ijbc} with the same training configuration.

\subsection{Exploratory Experiments}

{\bf Preprocessing and training setting.} We use CASIA-WebFace~\cite{WebFace} as training data and ResNet-50 as the backbone network architecture. Here WebFace~\cite{WebFace} dataset is cleaned and contains about $450$k facial images. RSA \cite{liu_2017_rsa} is adopted to images to extract facial areas and then aligns the faces using similarity transformation. All images are resized to $144\times 144$. Also, we conduct pixel value normalization by subtracting $127.5$ and then dividing by $128$. For all exploratory experiments, the size of a mini-batch is $512$ in every iteration.

\begin{figure*}
\begin{center}
   \includegraphics[width=1\linewidth]{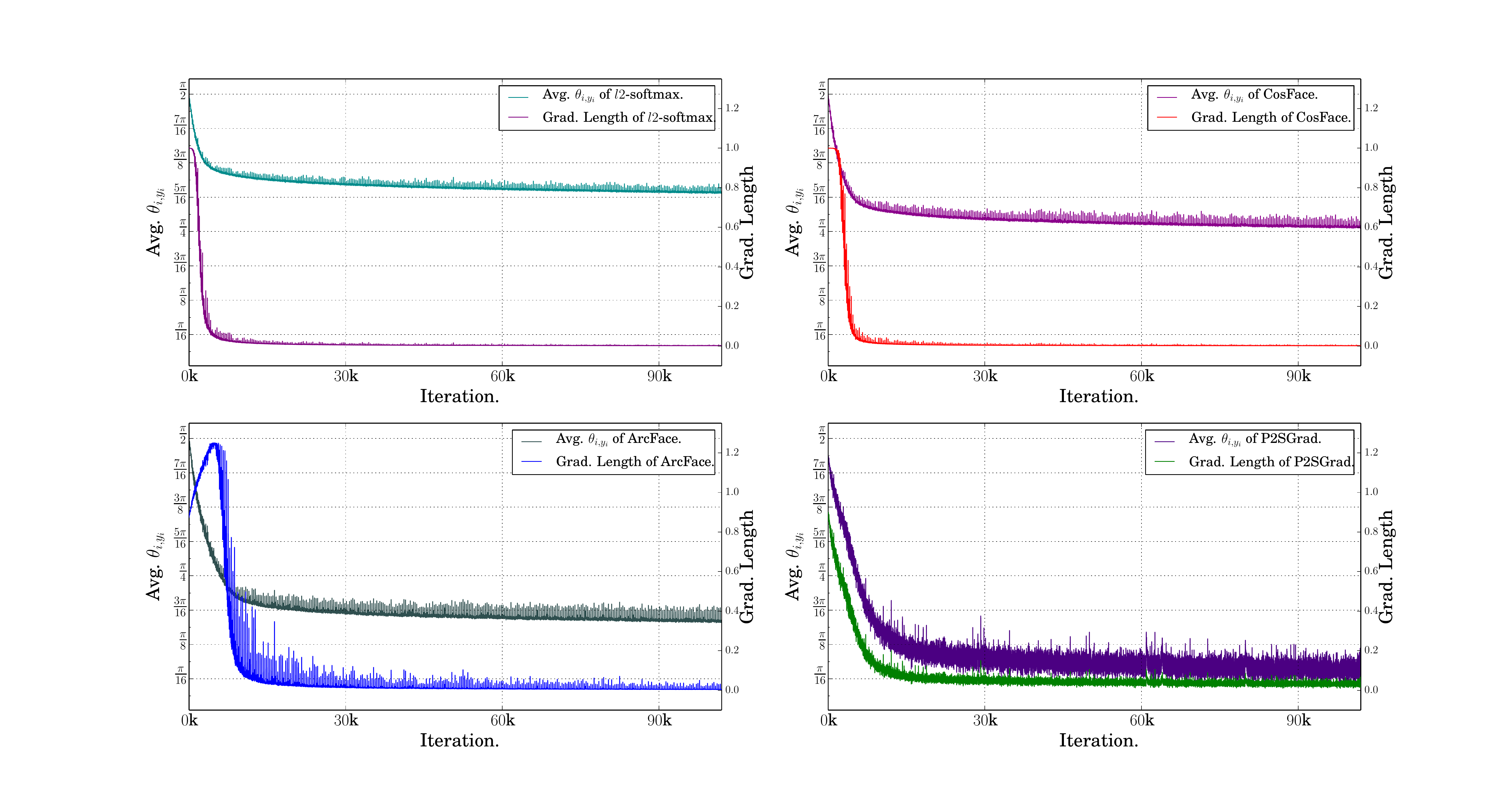}
\end{center}
   \caption{ 
        {Curves of $\theta_{i,y_i}$ and gradient lengths w.r.t. iteration.} Gradient lengths in existing cosine-based softmax losses (top-left, top-right, bottom-left) rapidly decrease to nearly $0$ while gradient length produced by P2SGrad (bottom-right) can match $\theta_{i,y_i}$ between $x_i$ and its ground truth class $y_i$. Best viewed in color.
    }
\label{fig:theta_prob_mapping_for_diff_margin}

\end{figure*}

{\bf The change of gradient length and $\theta_{i,y_i}$ w.r.t. iteration.}
Since P2SGrad aims to set up a reasonable mapping from $\theta_{i,y_i}$ to the length of gradients, it is necessary to visualize such mapping. In order to demonstrate the advancement of P2SGrad, we plot mapping curves of several cosine-based softmax losses in Fig. \ref{fig:theta_prob_mapping_for_diff_margin}. This figure clearly shows that P2SGrad produces more optimal gradient length according to the change of $\theta_{i,y_i}$.

\begin{table}

\begin{center}
    \begin{tabular}{|c|c|c|c|c|}  
    \hline
    \multicolumn{1}{|c|}{\multirow{2}{*}{Init. LR}} & \multicolumn{4}{|c|}{Method} \\
    \cline{2-5}&NormFace&CosFace&ArcFace&P2SGrad\\
    \hline\hline
    \multicolumn{1}{|c|}{$10^{-1}$}&   \multicolumn{1}{|c|}{${\times}$} & \multicolumn{1}{|c|}{${\times}$} & \multicolumn{1}{|c|}{${\times}$} & \multicolumn{1}{|c|}{$\surd$}\\
    \multicolumn{1}{|c|}{$10^{-2}$}&   \multicolumn{1}{|c|}{${\surd}$} & \multicolumn{1}{|c|}{${\times}$} & \multicolumn{1}{|c|}{${\times}$} & \multicolumn{1}{|c|}{${\surd}$} \\
    \multicolumn{1}{|c|}{$10^{-3}$} & \multicolumn{1}{|c|}{${\surd}$} & \multicolumn{1}{|c|}{${\surd}$} & \multicolumn{1}{|c|}{${\surd}$} & \multicolumn{1}{|c|}{${\surd}$} \\
    \multicolumn{1}{|c|}{$10^{-4}$}&\multicolumn{1}{|c|}{${\surd}$}&\multicolumn{1}{|c|}{${\surd}$}&\multicolumn{1}{|c|}{${\surd}$}&\multicolumn{1}{|c|}{${\surd}$}\\  
\hline
    \end{tabular}  
\end{center}
\caption{{The sensitiveness of initial learning rates.} This table shows whether our P2SGrad and these cosine-based softmax loss are trainable under different initial learning rates.}
\label{table:trainable}

\end{table} 

{\bf Robustness of initial learning rates.}
An important problem of margin-based loss is that they are difficult to train with large learning rates. The implementation of L-softmax \cite{L-softmax} and A-softmax \cite{A-softmax} use extra hyperparameters to adjust the margin so that the models are trainable. Thus a small initial learning rate is important for properly training angular-margin-based softmax losses. In contrast, shown in Table.~\ref{table:trainable}, our proposed P2SGrad is stable with large learning rates.

\begin{figure}[t]
\begin{center}
   \includegraphics[width=1\linewidth]{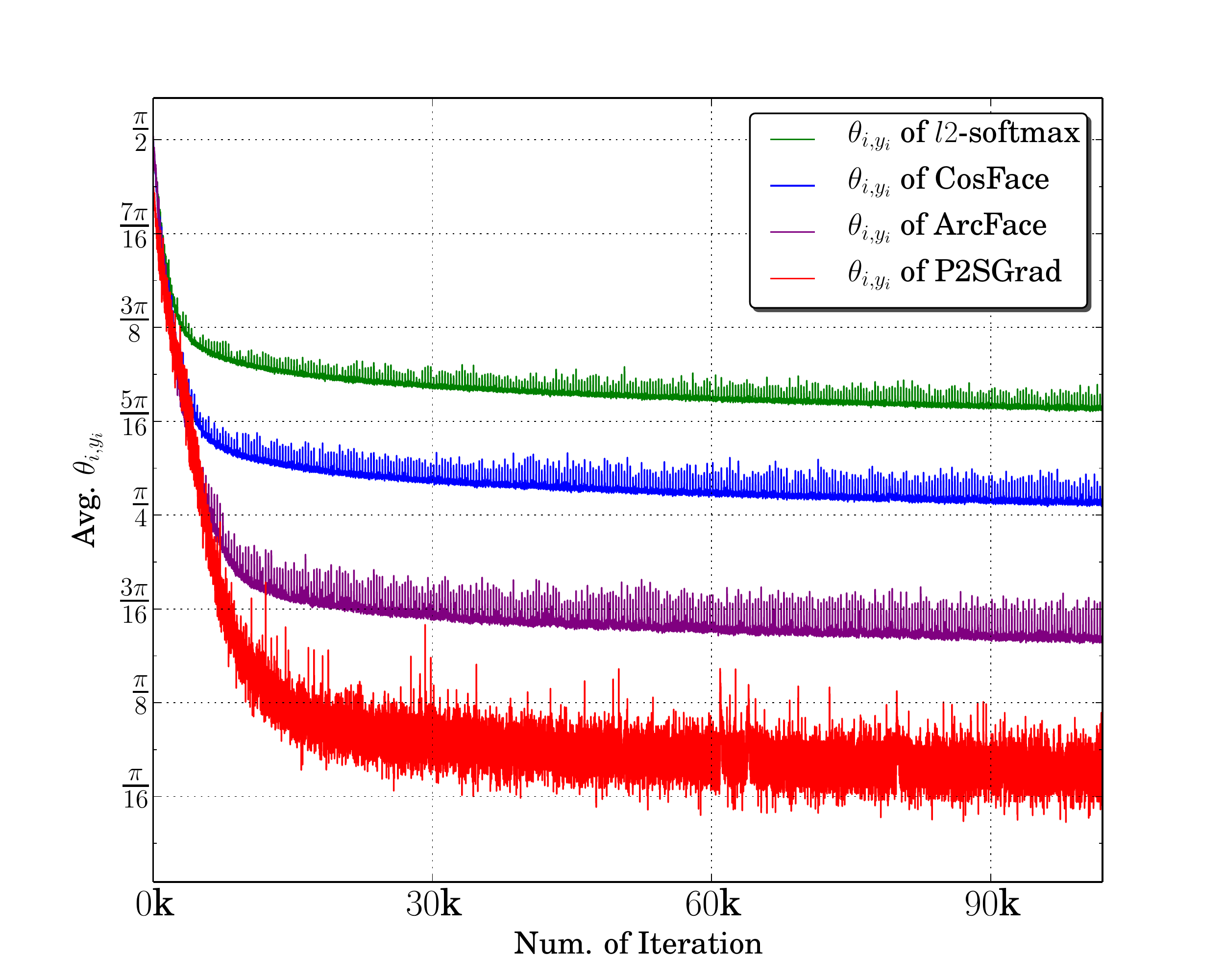}
\end{center}
   \caption{ 
        { The change of average $\theta_{i,y_i}$ w.r.t. iteration number.} $\theta_{i,y_i}$ represents the angle between $x_i$ and the weight vector of its ground truth class $y_i$. Curves by the proposed P2SGrad, $l2$-softmax loss \cite{L2-softmax}, CosFace \cite{CosFace} and ArcFace \cite{ArcFace} are shown.
    }
\label{fig:theta_iteration}

\end{figure}

\begin{table}
\begin{center}
\begin{tabular}{|l|c|c|c|}
\hline
\multirow{2}{*}{Method} & \multicolumn{3}{c|}{Num. of Iteration}\\
\cline{2-4}& $30$k & $60$k & $90$k \\
\hline\hline
$l2$-softmax~\cite{L2-softmax} & $81.50$ &$91.27$ & $97.92$\\
CosFace~\cite{CosFace}         & $83.63$ &$93.58$ & $99.05$\\
ArcFace~\cite{ArcFace}         & $85.32$ &$94.77$ & $99.47$\\
\hline
\bf{P2SGrad} & $\textbf{91.25}$ &$\textbf{97.38}$ & $\textbf{99.82}$\\
\hline
\end{tabular}
\end{center}
\caption{{Convergence rates of P2SGrad and compared losses.} With the same number of iterations, P2SGrad leads to the best performance.}
\label{tab:lfw_speed}

\end{table}

\begin{table*}
\begin{center}
\begin{tabular}{|c||c|c|c|c|c|c|}
\hline
    \multirow{2}{*}{Method} & \multicolumn{6}{c|}{Size of MegaFace Distractor} \\
\cline{2-7}&$10^1$&$10^2$&$10^3$&$10^4$&$10^5$&$10^6$\\
    \hline  \hline
  $l2$-softmax~\cite{L2-softmax}& $99.73\%$ & $99.49\%$ & $99.03\%$ & $97.85\%$ & $95.56\%$ & $92.05\%$\\
    CosFace~\cite{CosFace}& $99.82\%$ & $99.68\%$ & $99.46\%$ & $98.57\%$ & $97.58\%$ & $95.50\%$\\  
    ArcFace~\cite{ArcFace}   & $99.78\%$ & $99.65\%$ & $99.48\%$ & $98.87\%$ & $98.03\%$ & $96.88\%$\\
    \hline
    {\bf P2SGrad} & $\textbf{99.86\%}$ & $\textbf{99.70\%}$ & $\textbf{99.52\%}$ & $\textbf{98.92\%}$ & $\textbf{98.35\%}$ & $\textbf{97.25\%}$ \\
\hline
\end{tabular}
\end{center}

\caption{{Recognition accuracy on MegaFace.} Inception-ResNet~\cite{IR} models trained with different compared softmax loss and the same cleaned WebFace~\cite{WebFace} and MS1M~\cite{MS-Celeb-1M} training data.}
\label{tab:megaface_benchmark}

\end{table*}

\begin{table*}
\begin{center}
\begin{tabular}{|c||c|c|c|c|c|c|c|}
\hline
    \multirow{2}{*}{Method} & \multicolumn{7}{c|}{True Acceptance Rate @ False Acceptance Rate} \\
\cline{2-8}&$10^{-1}$&$10^{-2}$&$10^{-3}$&$10^{-4}$&$10^{-5}$&$10^{-6}$&$10^{-7}$\\
    \hline\hline  
VggFace \cite{parkhi2015deep}& $95.64\%$ & $87.13\%$ & $74.79\%$ & $59.75\%$ & $43.69\%$ & $32.20\%$&-\\

Crystal Loss \cite{ranjan2018crystal}& $99.06\%$ & $97.66\%$ & $95.63\%$ & $92.29\%$ & $87.35\%$ & $81.15\%$&$71.37\%$\\
    \hline
    
$l2$-softmax~\cite{L2-softmax}& $98.40\%$ & $96.45\%$ & $92.78\%$ & $86.33\%$ & $77.25\%$ & $62.61\%$ & $26.67\%$\\

CosFace \cite{CosFace}& $99.01\%$ & $97.55\%$ & $95.37\%$ & $91.82\%$ & $86.94\%$ & $76.25\%$ & $61.72\%$\\
    
ArcFace \cite{ArcFace}& $\textbf{99.07\%}$ & $97.75\%$ & $95.55\%$ & $92.13\%$ & $87.28\%$ & $82.15\%$ & $72.28\%$\\

    \hline
    {\bf P2SGrad} & $99.03\%$ & $\textbf{97.79\%}$ & $\textbf{95.58\%}$ & $\textbf{92.25\%}$ & $\textbf{87.84\%}$ & $\textbf{82.44\%}$ & $\textbf{73.16\%}$ \\
\hline
\end{tabular}
\end{center}
\caption{ {TARs by different compared softmax losses on the IJB-C 1:1 verification task.} The same training data (WebFace \cite{WebFace} and MS1M \cite{MS-Celeb-1M}) and Inception-ResNet \cite{IR} networks are used. Results of VggFace \cite{parkhi2015deep} and Crystal Loss \cite{ranjan2018crystal} are from \cite{ranjan2018crystal}.}
\label{tab:ijbc11_benchmark}

\end{table*}

{\bf Convergence rate.} The convergence rate is important for evaluating optimization methods. We evaluated the trained model's performance on Labeled Faces in the Wild (LFW) dataset of several cosine-based softmax losses and our P2SGrad method at different training periods.  LFW dataset is an academic test set for unrestricted face verification. Its testing protocol contains about $13,000$ images of about $1,680$ identities. There are $3,000$ positive matches and the same number of negative matches. Table.~\ref{tab:lfw_speed} shows the results with the same training configuration while Fig.~\ref{fig:theta_iteration} shows the decrease of average $\theta_{i,y_i}$ with P2SGrad is more rapid than other losses. These results reveal that our proposed P2SGrad can optimize neural network much faster.

\subsection{Evaluation on MegaFace}
\label{ssec:evalu_megaface}

{\bf Preprocessing and training setting.}
Besides the mentioned WebFace~\cite{WebFace} dataset, we add another public training dataset, MS1M~\cite{MS-Celeb-1M}, which contains about $2.35$M cleaned and aligned images. Here we use Inception-ResNet \cite{DR,IR} with a batch size of $512$ for training.

{\bf Evaluation results.}
MegaFace 1 million Challenge~\cite{MegaFace2} is a public identification benchmark to test the performance of facial identification algorithms. The distractor in MegaFace contains about $1,000,000$ images. Here we follow the cleaned testing protocol in \cite{ArcFace}. The results of P2SGrad on MegaFace dataset are shown in Table~\ref{tab:megaface_benchmark}. P2SGrad exceeds other compared cosine-based losses on MegaFace 1 million challenge with every size of distractor.

\subsection{Evaluation on IJBC 1:1 verification}

{\bf Preprocessing and training setting.} Same as \ref{ssec:evalu_megaface}.

{\bf Evaluation results.}
The IJB-C dataset \cite{ijbc} contains about $3,500$ identities with a total of $31,334$ still facial images and $117,542$ unconstrained video frames. The entire IJB-C testing protocols are designed to test detection, identification, verification and clustering of faces. In the 1:1 verification protocol, there are $19,557$ positive matches and $15,638,932$ negative matches. Therefore, we test Ture Acceptance Rates at very strict False Acceptance Rates. Table.~\ref{tab:ijbc11_benchmark} exhibits that P2SGrad surpasses all other cosine-based losses.

\section{Conclusion}

we comprehensively discussed the limitations of the forward and backward processes in training deep model for face recognition. To deal with the limitations, we proposed a simple but effective gradient method, P2SGrad, which is hyperparameter free and leads to better optimization results. Unlike previous methods which focused on loss functions, we improve the deep network training by using carefully designed gradients. Extensive experiments validate the robustness and fast convergence of the proposed method. Moreover, experimental results show that P2SGrad achieves superior performance over state-of-the-art methods on several challenging face recognition benchmarks.

\textbf{Acknowledgements.} This work is supported in part by SenseTime Group Limited, in part by the General Research Fund through the Research Grants Council of Hong Kong under Grants CUHK14202217, CUHK14203118, CUHK14205615, CUHK14207814, CUHK14213616, CUHK14208417, CUHK14239816, in part by CUHK Direct Grant, and in part by National Natural Science Foundation of China (61472410) and the Joint Lab of CAS-HK.

{\small
\bibliographystyle{ieee}
\bibliography{P2SGrad}

\begin{thebibliography}{10}\itemsep=-1pt

\bibitem{belhumeur1997eigenfaces}
Peter~N. Belhumeur, Jo{\~a}o~P Hespanha, and David~J. Kriegman.
\newblock Eigenfaces vs. fisherfaces: Recognition using class specific linear
  projection.
\newblock {\em IEEE Transactions on pattern analysis and machine intelligence},
  19(7):711--720, 1997.

\bibitem{contrastiveloss}
S Chopra, R Hadsell, and Y Lecun.
\newblock Learning a similarity metric discriminatively, with application to
  face verification.
\newblock In {\em Computer Vision and Pattern Recognition, 2005. CVPR 2005.
  IEEE Computer Society Conference on}, pages 539--546 vol. 1, 2005.

\bibitem{ArcFace}
Jiankang Deng, Jia Guo, and Stefanos Zafeiriou.
\newblock Arcface: Additive angular margin loss for deep face recognition.
\newblock {\em arXiv preprint arXiv:1801.07698}, 2018.

\bibitem{MS-Celeb-1M}
Yandong Guo, Lei Zhang, Yuxiao Hu, Xiaodong He, and Jianfeng Gao.
\newblock Ms-celeb-1m: A dataset and benchmark for large-scale face
  recognition.
\newblock In {\em European Conference on Computer Vision}, pages 87--102.
  Springer, 2016.

\bibitem{DR}
Kaiming He, Xiangyu Zhang, Shaoqing Ren, and Jian Sun.
\newblock Deep residual learning for image recognition.
\newblock In {\em Proceedings of the IEEE conference on computer vision and
  pattern recognition}, pages 770--778, 2016.

\bibitem{hu2017squeeze}
Jie Hu, Li Shen, and Gang Sun.
\newblock Squeeze-and-excitation networks.
\newblock {\em arXiv preprint arXiv:1709.01507}, 2017.

\bibitem{LFW}
Gary~B Huang, Manu Ramesh, Tamara Berg, and Erik Learned-Miller.
\newblock Labeled faces in the wild: A database for studying face recognition
  in unconstrained environments.
\newblock Technical report, Technical Report 07-49, University of
  Massachusetts, Amherst, 2007.

\bibitem{MegaFace2}
Ira Kemelmacher-Shlizerman, Steven~M Seitz, Daniel Miller, and Evan Brossard.
\newblock The megaface benchmark: 1 million faces for recognition at scale.
\newblock In {\em Proceedings of the IEEE Conference on Computer Vision and
  Pattern Recognition}, pages 4873--4882, 2016.

\bibitem{krizhevsky2012imagenet}
Alex Krizhevsky, Ilya Sutskever, and Geoffrey~E Hinton.
\newblock Imagenet classification with deep convolutional neural networks.
\newblock In {\em Advances in neural information processing systems}, pages
  1097--1105, 2012.

\bibitem{Pubfig}
Neeraj Kumar, Alexander~C Berg, Peter~N Belhumeur, and Shree~K Nayar.
\newblock Attribute and simile classifiers for face verification.
\newblock In {\em Computer Vision, 2009 IEEE 12th International Conference on},
  pages 365--372. IEEE, 2009.

\bibitem{A-softmax}
Weiyang Liu, Yandong Wen, Zhiding Yu, Ming Li, Bhiksha Raj, and Le Song.
\newblock Sphereface: Deep hypersphere embedding for face recognition.
\newblock In {\em The IEEE Conference on Computer Vision and Pattern
  Recognition (CVPR)}, volume~1, 2017.

\bibitem{L-softmax}
Weiyang Liu, Yandong Wen, Zhiding Yu, and Meng Yang.
\newblock Large-margin softmax loss for convolutional neural networks.
\newblock In {\em ICML}, pages 507--516, 2016.

\bibitem{liu_2017_coco_v1}
Yu Liu, Hongyang Li, and Xiaogang Wang.
\newblock Learning deep features via congenerous cosine loss for person
  recognition.
\newblock {\em arXiv preprint arXiv:1702.06890}, 2017.

\bibitem{liu_2017_coco_v2}
Yu Liu, Hongyang Li, and Xiaogang Wang.
\newblock Rethinking feature discrimination and polymerization for large-scale
  recognition.
\newblock {\em arXiv preprint arXiv:1710.00870}, 2017.

\bibitem{liu_2017_rsa}
Yu Liu, Hongyang Li, Junjie Yan, Fangyin Wei, Xiaogang Wang, and Xiaoou Tang.
\newblock Recurrent scale approximation for object detection in cnn.
\newblock In {\em IEEE International Conference on Computer Vision}, 2017.

\bibitem{ijbc}
Brianna Maze, Jocelyn Adams, James~A Duncan, Nathan Kalka, Tim Miller, Charles
  Otto, Anil~K Jain, W~Tyler Niggel, Janet Anderson, Jordan Cheney, et~al.
\newblock Iarpa janus benchmark--c: Face dataset and protocol.
\newblock In {\em 11th IAPR International Conference on Biometrics}, 2018.

\bibitem{MegaFace1}
Aaron Nech and Ira Kemelmacher-Shlizerman.
\newblock Level playing field for million scale face recognition.
\newblock In {\em 2017 IEEE Conference on Computer Vision and Pattern
  Recognition (CVPR)}, pages 3406--3415. IEEE, 2017.

\bibitem{parkhi2015deep}
Omkar~M Parkhi, Andrea Vedaldi, Andrew Zisserman, et~al.
\newblock Deep face recognition.
\newblock In {\em BMVC}, volume~1, page~6, 2015.

\bibitem{ranjan2018crystal}
Rajeev Ranjan, Ankan Bansal, Hongyu Xu, Swami Sankaranarayanan, Jun-Cheng Chen,
  Carlos~D Castillo, and Rama Chellappa.
\newblock Crystal loss and quality pooling for unconstrained face verification
  and recognition.
\newblock {\em arXiv preprint arXiv:1804.01159}, 2018.

\bibitem{L2-softmax}
Rajeev Ranjan, Carlos~D Castillo, and Rama Chellappa.
\newblock L2-constrained softmax loss for discriminative face verification.
\newblock {\em arXiv preprint arXiv:1703.09507}, 2017.

\bibitem{FaceNet}
Florian Schroff, Dmitry Kalenichenko, and James Philbin.
\newblock Facenet: A unified embedding for face recognition and clustering.
\newblock In {\em Proceedings of the IEEE conference on computer vision and
  pattern recognition}, pages 815--823, 2015.

\bibitem{DeepID2}
Yi Sun, Yuheng Chen, Xiaogang Wang, and Xiaoou Tang.
\newblock Deep learning face representation by joint
  identification-verification.
\newblock In {\em Advances in neural information processing systems}, pages
  1988--1996, 2014.

\bibitem{DeepID3}
Yi Sun, Ding Liang, Xiaogang Wang, and Xiaoou Tang.
\newblock Deepid3: Face recognition with very deep neural networks.
\newblock {\em arXiv preprint arXiv:1502.00873}, 2015.

\bibitem{IR}
Christian Szegedy, Sergey Ioffe, Vincent Vanhoucke, and Alexander~A Alemi.
\newblock Inception-v4, inception-resnet and the impact of residual connections
  on learning.
\newblock In {\em AAAI}, volume~4, page~12, 2017.

\bibitem{szegedy2015going}
Christian Szegedy, Wei Liu, Yangqing Jia, Pierre Sermanet, Scott Reed, Dragomir
  Anguelov, Dumitru Erhan, Vincent Vanhoucke, Andrew Rabinovich, et~al.
\newblock Going deeper with convolutions.
\newblock In {\em CVPR}, 2015.

\bibitem{DeepFace}
Yaniv Taigman, Ming Yang, Marc'Aurelio Ranzato, and Lior Wolf.
\newblock Deepface: Closing the gap to human-level performance in face
  verification.
\newblock In {\em Proceedings of the IEEE conference on computer vision and
  pattern recognition}, pages 1701--1708, 2014.

\bibitem{AM-softmax}
Feng Wang, Weiyang Liu, Haijun Liu, and Jian Cheng.
\newblock Additive margin softmax for face verification.
\newblock {\em arXiv preprint arXiv:1801.05599}, 2018.

\bibitem{NormFace}
Feng Wang, Xiang Xiang, Jian Cheng, and Alan~L Yuille.
\newblock Normface: $ l\_2 $ hypersphere embedding for face verification.
\newblock {\em arXiv preprint arXiv:1704.06369}, 2017.

\bibitem{CosFace}
Hao Wang, Yitong Wang, Zheng Zhou, Xing Ji, Zhifeng Li, Dihong Gong, Jingchao
  Zhou, and Wei Liu.
\newblock Cosface: Large margin cosine loss for deep face recognition.
\newblock {\em arXiv preprint arXiv:1801.09414}, 2018.

\bibitem{tripletloss2}
Kilian~Q Weinberger and Lawrence~K Saul.
\newblock Distance metric learning for large margin nearest neighbor
  classification.
\newblock {\em Journal of Machine Learning Research}, 10(Feb):207--244, 2009.

\bibitem{centerloss}
Yandong Wen, Kaipeng Zhang, Zhifeng Li, and Yu Qiao.
\newblock A discriminative feature learning approach for deep face recognition.
\newblock In {\em European Conference on Computer Vision}, pages 499--515.
  Springer, 2016.

\bibitem{WebFace}
Dong Yi, Zhen Lei, Shengcai Liao, and Stan~Z Li.
\newblock Learning face representation from scratch.
\newblock {\em arXiv preprint arXiv:1411.7923}, 2014.

\bibitem{rangeloss}
Xiao Zhang, Zhiyuan Fang, Yandong Wen, Zhifeng Li, and Yu Qiao.
\newblock Range loss for deep face recognition with long-tailed training data.
\newblock In {\em Proceedings of the IEEE Conference on Computer Vision and
  Pattern Recognition}, pages 5409--5418, 2017.

\end{thebibliography}
}

\end{document}